\documentclass[letterpaper, 10 pt, conference]{ieeeconf}
\IEEEoverridecommandlockouts   
\usepackage{lipsum} 
\usepackage[T1]{fontenc}
\usepackage{graphicx}
\usepackage{graphics}
\usepackage{multirow}
\usepackage{float}
\usepackage{hyperref}
\usepackage{epstopdf}
\usepackage{enumerate}
\usepackage{multicol,multirow}
\usepackage{booktabs}
\usepackage{cite}
\usepackage{amsmath}
\usepackage{academicons}
\usepackage{amssymb}
\usepackage{amsfonts}
\usepackage{bbm}
\usepackage{balance}
\usepackage{paralist}
\usepackage[binary-units=true]{siunitx}
\graphicspath{{figures/}}
\usepackage[ruled,vlined,linesnumbered]{algorithm2e}
\let\oldnl\nl
\newcommand{\nonl}{\renewcommand{\nl}{\let\nl\oldnl}}

\graphicspath{{figures/}} 

\begin{document}
\thispagestyle{empty}
\fbox{
\parbox{\textwidth}{
© 2021 IEEE. Personal use of this material is permitted.  Permission from IEEE must be obtained for all other uses, in any current or future media, including reprinting/republishing this material for advertising or promotional purposes, creating new collective works, for resale or redistribution to servers or lists, or reuse of any copyrighted component of this work in other works.}}
\newpage
	\title{\LARGE\bf
	    A Passive Navigation Planning Algorithm for Collision-free Control of Mobile Robots
	}

	\author{Carlo Tiseo, Vladimir Ivan, Wolfgang Merkt, Ioannis Havoutis, Michael Mistry and Sethu Vijayakumar

		\thanks{Carlo Tiseo, Vladimir Ivan, Michael Mistry and Sethu Vijayakumar are with the Edinburgh Centre for Robotics, Institute of Perception Action and Behaviour, School of Informatics, University of Edinburgh. Wolfgang Merkt and Ioannis Havoutis are with the Oxford Robotics Institute, University of Oxford. Email: \texttt{carlo.tiseo@ed.ac.uk}}
	   \thanks{This work has been supported by the following grants: EPSRC UK RAI Hubs ORCA (EP/R026173/1), NCNR (EPR02572X/1) and FAIR-Space (EP/R026092/1); and EU Horizon 2020 projects MEMMO (780684) and THING (ICT-2017-1).}
	}
\maketitle
	
\begin{abstract}
Path planning and collision avoidance are challenging in complex and highly variable environments due to the limited horizon of events. In literature, there are multiple model- and learning-based approaches that require significant computational resources to be effectively deployed and they may have limited generality. We propose a planning algorithm based on a globally stable passive controller that can plan smooth trajectories using limited computational resources in challenging environmental conditions. The architecture combines the recently proposed fractal impedance controller with elastic bands and regions of finite time invariance. As the method is based on an impedance controller, it can also be used directly as a force/torque controller. We validated our method in simulation to analyse the ability of interactive navigation in challenging concave domains via the issuing of via-points, and its robustness to low bandwidth feedback. A swarm simulation using 11 agents validated the scalability of the proposed method. We have performed hardware experiments on a holonomic wheeled platform validating smoothness and robustness of interaction with dynamic agents (i.e., humans and robots). The computational complexity of the proposed local planner enables deployment with low-power micro-controllers lowering the energy consumption compared to other methods that rely upon numeric optimisation.
\end{abstract}

\IEEEpeerreviewmaketitle

\section{Introduction}
Robust path/navigation planning and control for mobile robots is essential to deploy robots in extreme environments to act as a medium for remote inspection and intervention in industrial and search and rescue scenarios. Successful deployment of robots in such applications requires robustness and manoeuvrability to be controlled to perform dexterous tasks and accurate navigation \cite{xin2020,xin2020CASE,angelini2019,merkt2017robust}. A major challenge in these environments is the inherently short horizon of events, the lack of accurate environmental models and expected perturbations \cite{majumdar2013, Llamazares2013}. These limitations make it extremely difficult to rely on complex model-based control and planning architectures due to long computational time and their need for accurate state estimation. On the other hand, reactive control architectures are usually sufficiently robust in these environments, but they are generally unable to provide the dexterity required to operate in such conditions \cite{stecz2020,liu2019,li2020,qu2020,merkt2019}. There has been a long-standing problem in robotics on how to integrate these characteristics in a single architecture, which provides the ability to perform complex manoeuvring without renouncing the robustness of reactive controllers \cite{majumdar2013,quinlan1993, brock2000}.

\begin{figure}[t]
\centering
\includegraphics[width=\columnwidth, trim=0cm 1.0cm 0cm 1.0cm, clip]{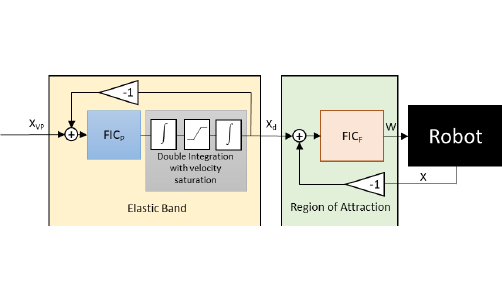}
\caption{The proposed architecture takes as input the coordinates of the next via-point $X_{\text{VP}}$ to the Elastic Band module. The Fractal Impedance Controller used for the planning FIC$_{\text{P}}$ converts the position error into acceleration which is later provided to a double integrator with saturation of the velocities. The integrator output $X_\text{d}$ is then provided as input to the Fractal Impedance Controller FIC$_\text{F}$ that defines the region of attraction and, consequently, the region of finite-time invariance, and outputs the control wrench ($W$).}
\label{fig:CtrlArc}
\end{figure}   

Robot navigation algorithms and architectures are divided into global and local planning. 
Common local planners use dynamic window or time-elastic band approaches for local collision avoidance \cite{fox1997dynamic,rosmann2017integrated}. Recent advances in path planning and collision avoidance algorithms improved the manoeuvrability of remotely operated and autonomous robots significantly. These methods are often isolated from the control architecture, and they require to encode the robot dynamics, environment and the interaction with the robot as optimisation constraints \cite{stecz2020,rosmann2017integrated,merkt2019,zhou2019,qu2020,wen2015}. A major implication of this approach is that every time one of the conditions considered in the optimisation problem changes, it invalidates the computed solution. Although currently available computers have drastically reduced the computational times required to solve the problem, they are still heavily affected by the curse of dimensionality \cite{stecz2020,zhou2019,qu2020}. Furthermore, the results demonstrating that these methods can be performed online are often obtained with power CPUs, which are not ideal to be deployed on battery-operated systems \cite{rosmann2017integrated,zhou2019}. In summary, path planning algorithms are greatly affected by the environment, and an increase in its complexity drastically increases the energetic cost. This may jeopardise the ability to find a solution online. A possible solution for the curse of dimensionality can be obtained by decomposing the problem into local problems where the problem dynamics can be considered time-invariant in the neighbourhood of a via-point (or knot point) \cite{majumdar2013}. However, this approach requires contiguous regions of finite-time invariance to overlap in order to guarantee the existence of a solution to the problem \cite{majumdar2013}. Therefore, an architecture based on such an approach still requires the solutions of a higher-level optimisation problem to verify the overlap, unless it is possible to guarantee a global region of attraction for each via-point, allowing to decouple the global strategy from local stability. 

Our earlier work \cite{tiseo2018thesis} theorised the possibility of developing a bio-inspired planner based on an integration between the passive motion paradigm and elastic bands, represented in \autoref{fig:CtrlArc}. The passive motion paradigm is a planner based on an impedance controller that generates a basin of attraction around the desired state \cite{tommasino2017}. Meanwhile, elastic bands and bubbles are a classical approach to algorithms for path planning and obstacle avoidance in unstructured environments \cite{brock2000,quinlan1993}, and they are still deployed in state-of-the-art drone navigation algorithms in unstructured environments \cite{zhou2019}. However, as previously mentioned, the successful implementation of this method requires the availability of a controller to generate a region of finite-time invariance around the current robot state. Here, the region of finite-time invariance is defined as the region of attraction around a fix-point in which the system dynamics can be defined as time-invariant \cite{majumdar2013}. Traditional passive controllers are not well suited for this application because they rely on numerical integration to passivise the controller's damping, which makes compromises on their performances, especially in highly dynamic environments \cite{babarahmati2019,tiseo2020}. However, the recently introduced Fractal Impedance Controller (FIC) \cite{babarahmati2019,babarahmati2020,tiseo2020bio,tiseo2020}, a non-linear passive controller, can be exploited to take advantage of the autonomous trajectories of the fractal attractor to achieve accurate trajectory tracking and robustness to unknown disturbance.

This manuscript integrates the fractal impedance controller with the passive motion paradigm, proposing a computationally inexpensive control architecture that can solve the elastic band optimisation in real-time without the need of running a numerical optimisation.

\section{Method}
The proposed planning architecture (\autoref{fig:CtrlArc}) consists of a cascade of two controllers. The first controller (FIC$_\text{P}$) acts as an elastic band pulling the robot towards the via-point (X$_\text{VP}$). The controller output is then transformed into the next desired state (X$_\text{d}$) using the forward dynamics defined by impedance controller. X$_\text{d}$ is the input of the second controller (FIC$_\text{F}$) that generates the region of attraction surrounding the desired state. This controller deals with the environmental interaction ensuring that the robot converges to the desired state with stable, smooth trajectories.

\subsection{Intrinsic Smoothness and Stability Properties of the FIC}
\begin{figure}[!htbp]
\centering
\includegraphics[width=0.9\columnwidth, trim=4cm 10.5cm 4cm 10.5cm, clip]{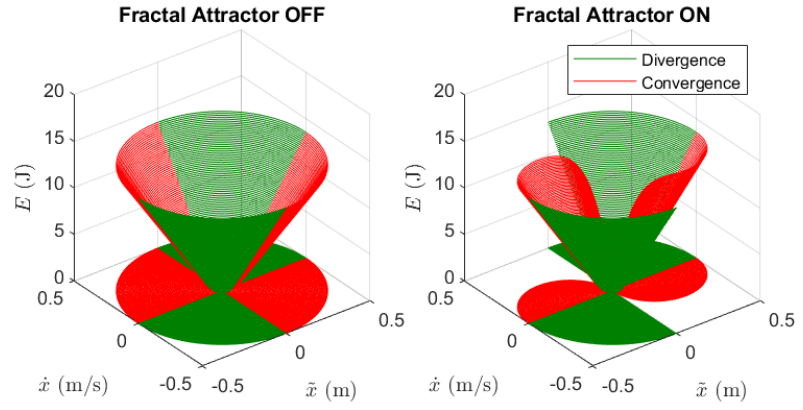}
\caption{Effect of the fractal attractor on an impedance controller without damping, showing how the algorithm always ensures a critically damped behaviour. The autonomous trajectories of the fractal attractor during divergence and convergence in phase space show how the attractor topology scales with the energy accumulated in the controller but it is not deformed. Trajectories for deformations greater than \SI{0.5}{\meter} are omitted from the image.}
\label{fig:Attractor}
\end{figure}

The Fractal Impedance Controller (FIC) can be used to generate attractor landscapes (see e.g., as shown in \autoref{fig:Attractor}) around the desired pose. The attractor's geodesics scale with the controller's potential energy without altering their topology (\autoref{fig:Attractor}) ensuring that autonomous trajectories are harmonic functions. The fractal impedance controller is a passive/conservative system, which implies that it can be superimposed along with other controllers without compromising the system stability \cite{tiseo2020}. This property allows building hierarchical architectures of semi-autonomous controllers, which easily scale in coordinated multi-agents application (e.g., swarms) since the controllers' stability are independent from each other.

This type of controller has been deployed in controlling 7-DoF manipulators and haptic devices in both impedance and admittance architectures \cite{babarahmati2019,babarahmati2020,tiseo2020,tiseo2020bio}. We experimentally proved that independent controllers can be stacked together to accurately control a redundant manipulator without relying on the inverse dynamics and compromising the arm stability \cite{tiseo2020}. The equation for a mono-dimensional fractal impedance controller introduced in \cite{tiseo2020bio} is:
\begin{equation} 
h_{\text{e}}=
     \left\{\begin{array}{ll}
            F(\tilde{x}),& \text{~Divergence}\\
            \frac{F(\tilde{x}_{M})}{\tilde{x}_{\text{M}}}(2\tilde{x}-\tilde{x}_{\text{M}}),&\text{~Convergence}
       \end{array}\right.
 \label{alg1}
\end{equation}
\noindent where $F(\tilde{x})$ is the desired force/torque profile, $h_e$ the end-effector force/torque, $x_d$ the desired state, $x$ the current state, $\tilde{x}=x_d-x$ is the state error, and $X_\text{m}$ the maximum displacement reached during the divergence phase. 

This latest formulation of the FIC \cite{tiseo2020bio} introduces the concept of a virtual antagonist actuator that smooths the force profile transition when switching from divergence to convergence. The controller stability is further independent from the force profile, which can also be updated online as long as the force remains finite without non-removable discontinuities \cite{babarahmati2019,tiseo2020}.

\subsection{Planner and Control Architectures}
As mentioned before, the proposed architecture is composed of two controllers as shown in \autoref{fig:CtrlArc}. FIC$_\text{P}$ generates the smooth trajectory towards the desired state and FIC$_\text{F}$ determines a region of attraction around its output to deal with the environmental interaction.

\subsubsection{Elastic Band}
Elastic bands have been introduced in the early nineties for adapting a planned path to moving obstacles in mobile robots \cite{quinlan1993}. They are founded on the principle that a curve can be represented as an elastic band deformed by the encountered obstacles. Concurrently, the paper also defined a bubble---a potential energy field---surrounding the robot that was used to keep the robot away from unknown obstacles by acting as a virtual force that deformed the elastic band. The framework was later refined to be also applied to articulated objects mounted on mobile robots, which integrates it with virtual potential fields pushing the robot away from obstacles \cite{brock2000}. The framework is still in used in modern path planning algorithms such as \cite{zhou2019} where it is deployed for aggressive drone flight path planning with outstanding performance. However, all these approaches solve optimisation algorithms where constraints are required to obtain smooth trajectories compatible with the system dynamics capabilities. 

The proposed method takes advantage of the properties of the fractal impedance controller to formulate the elastic band optimisation as a Model Predictive Controller (MPC) projecting the information to the next time step which ensures smooth harmonic trajectories and global asymptotic stability \cite{tiseo2020bio}. The smoothness is guaranteed by the autonomous $C^\infty$ trajectories of the attractor,  which being harmonics makes it a necessary and sufficient condition for generating differentiable manifolds (i.e. smooth) \cite{ji2017riemann}. The asymptotic stability is guaranteed from the encoding of the hardware capabilities in the controller \cite{babarahmati2019}. The combination of these properties allows to use the controller for reactive trajectory planning. It further ensures better robustness to unknown perturbations because, differently from other methods, it does not require \textit{a priori} knowledge of the environment to validate the numerical stability of the optimisation.

For FIC$_\text{P}$, we use the following force profile:
\begin{equation}
\label{Eq:FICP1}
	    F_\text{P}(\tilde{x})=\begin{cases}
	    K_\text{d}(x_\text{VP}-x_\text{d}),~\text{  F}<\text{F}_\text{M}\\
	    F_\text{M}=M_\text{d}a_\text{M},~~~\text{Otherwise}
	    \end{cases}
\end{equation}
\noindent where $K_\text{P}$ is the spring constant of the elastic band, $x_\text{VP}$ is the input via-point, $x_\text{d}$ is the output provided to the FIC$_\text{F}$, $a_\text{M}$ is the maximum acceleration achievable with the robot, and $M_\text{d}$ is the desired apparent inertia for the robot that has to be greater than or equal to the robot mass. The output force is then divided by the total mass M$_\text{d}$ and integrated twice to obtain the desired target configuration $X_\text{d}$. Furthermore, a saturation in the velocity is added prior to the second integrator to enforce desired maximum velocities without limiting it through the robot acceleration.

\subsubsection{Region of Attraction as Regions of Finite-Time Invariance}
Majumdar and Tedrake~\cite{majumdar2013} proposed to use the region of attraction for path planning for nonlinear dynamical systems in highly variable environmental conditions. The region of attraction is defined as the states surrounding a stable point where the system can be stabilised. They proposed that the region of attraction for a time-varying system (e.g., an UAV flying in a forest) can be considered as a region of finite-time invariance. This assumption enables their method to divide the path in regions where the dynamics of the system can be considered invariant, simplifying the optimisation problem. These regions are funnels that can be combined together to solve a path planning optimisation, which guarantees the existence of a stable path as long as adjacent regions overlap \cite{burridge1999sequential,majumdar2013}.

The stability properties of the fractal impedance controller guarantee that these regions always overlap, removing the need for a numerical optimisation to verify the movement feasibility. Therefore, the combination of these two frameworks guarantees that it is always possible to plan stable trajectories between two points as long as the external perturbation is compatible with the physical limits of the robot. The region of attraction is implemented by the FIC$_\text{F}$ using the following force profile, based on \cite{tiseo2020bio}:
\begin{equation}
    \label{ForceFICf}
    F_{\text{F}}=\left\{
    \begin{array}{ll}
      K_0 (x_\text{d}-x)=K_0\tilde{x},  & |\tilde{x}|<\tilde{x}_{0}\\
 \text{sgn} (\tilde{x}) (\Delta F (1-e^{-\frac{|\tilde{x}| - \tilde{x}_0}{b}})+&\tilde{x}_{0}\le |\tilde{x}|~ \& \\
 + K_0 \tilde{x}_0),   & |\tilde{x}|<\tilde{x}_{b}\\
     \text{sgn} (\tilde{x})  F_{\text{Mf}},   & \text{Otherwise}
    \end{array}\right.
\end{equation}
\noindent where $K_0$ is the constant stiffness, $\tilde{x}_0$ is the position error at the end of the linear spring behaviour, $\tilde{x}_b$ is the position error when the force saturates to its maximum value ($F_\text{Mf}$), $\Delta F=F_\text{Mf}-K_0\tilde{x}_0$, and $b=(\tilde{x}_\text{b}-\tilde{x}_\text{0})/20$ is the saturation speed of the non-linear section of the force profile that reaches \SI{99.9}{\percent} before $\tilde{x}_\text{b}$.

\section{Experimental Validation}
Visualisations of the simulation and video of the hardware experiments are shown in the supplementary video that is also available at \url{https://youtu.be/KCP3QQpEAvE}. 

\subsection{Simulations}
\begin{figure}[t]
\centering
\includegraphics[width=0.9\columnwidth, trim=0.25cm 2.0cm 0.25cm -0.5cm, clip]{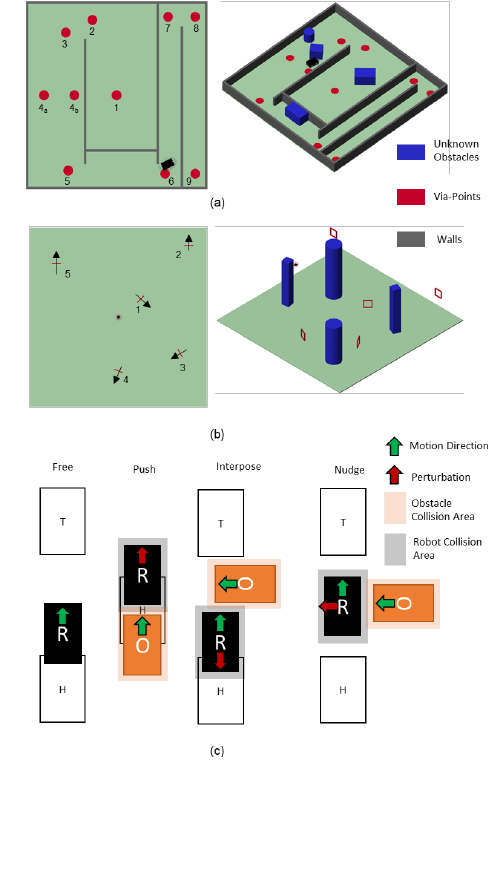}%
\caption{(a) The maze used for testing Ada robot and the 10 assigned via-points. The robot has to reach 9 and go back to 1. The path will initially pass by 4$_\text{a}$ and will go back though 4$_\text{b}$. (b) The circuit used for testing an UAV and the five gates the robot has to pass though in numerical order and in the direction indicated by the arrows. (c) The four tasks used in the experiment for testing the effect of interaction with an obstacle (O) when the robot (R) is either in the home position (H) or moving towards a target (T).}
\label{fig:ExpSetups}
\end{figure}

We validated our method in simulation using the Simulink Simscape library in Matlab2020a (Mathworks, USA). The simulations are performed on an Intel i7-7700HQ @ \SI{2.8}{\giga\hertz} with \SI{16}{\giga\byte} of RAM. We use the \texttt{ode4} solver with a sampling time of $10^{-3}~\si{\second}$. The simulations were performed for a holonomic mobile robot and an Unmanned Aerial Vehicle (UAV) in 3D mazes. We compare the performance in an empty maze and then add obstacles that the robot is unaware of, as shown in \autoref{fig:ExpSetups}. It shall be noted that all the simulations include an external viscous force field.

\subsubsection{Manoeuvrability and Robustness}
\label{AdaMazeR}
One of the main problems of numerical optimisation is ensuring the boundary conditions when passing through a knot point, which is essential to retain a differential manifold. These simulations are designed to validate that the proposed method guarantees \textit{a priori} that the planner energy (i.e., cost function) is a globally differentiable manifold (i.e., smooth) and, as a direct consequence, it is robust to numerical instability. It shall be noted that being autonomous trajectory of the proposed planner harmonic trajectory, this property is guaranteed by the Fourier series \cite{ji2017riemann}. We will also show that this property is maintained even in the presence of low-bandwidth feedback, that is usually a major challenge to numerical stability and introduces drift in state estimators \cite{li2020}.  We tested this on both a wheeled and a flying robot. The tracking Root Mean Square Errors (RMSE) for the simulations are included in the Appendix to show that the robots are pushed at their limits being their errors beyond $\tilde{x_b}$ (\autoref{ForceFICf}).

\begin{figure}[t]
\centering
\includegraphics[width=0.95\columnwidth, trim=5.5cm 9.5cm 5.7cm 9.5cm, clip]{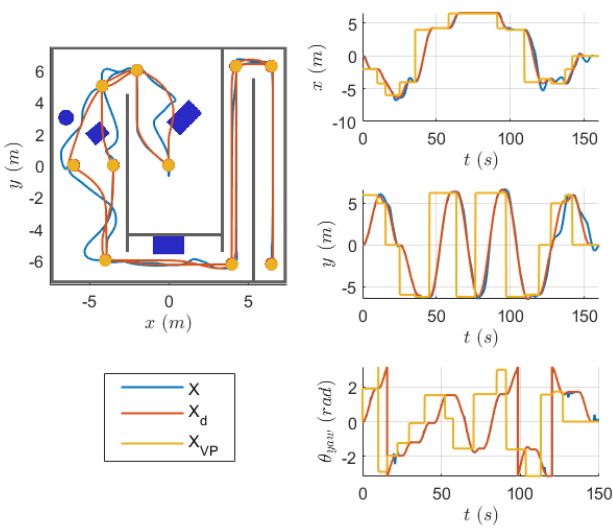}
\caption{Ada Robot simulations results with obstacles show that the proposed method can handle small unknown concavities. The spikes in the plots are due to the wrapping of the angles between $-\pi$ and $\pi$.}
\label{fig:AdaResults}
\end{figure}

The Ada500 robot is a high-performance omni-directional Autonomous Mobile Robot (AMR)  \cite{merkt2019towards}. The maximum linear acceleration is $0.5~\si{\meter~\second^{-2}}$, and the maximum angular acceleration is $0.92~\si{\radian~\second^{-2}}$. The results with unknown obstacles are shown in \autoref{fig:AdaResults}. The trajectories are smooth and respect the physical limitation of the robot (i.e., maximum velocity, maximum accelerations and maximum forces). The robot reached the terminal positions with negligible errors in all the degrees of freedom for all simulations, and passed closed enough to each via point to trigger the issuing of the subsequent via point.

\begin{figure}[t]
\centering
\includegraphics[width=0.95\columnwidth, trim=5.25cm 9.5cm 5.7cm 9.18cm, clip]{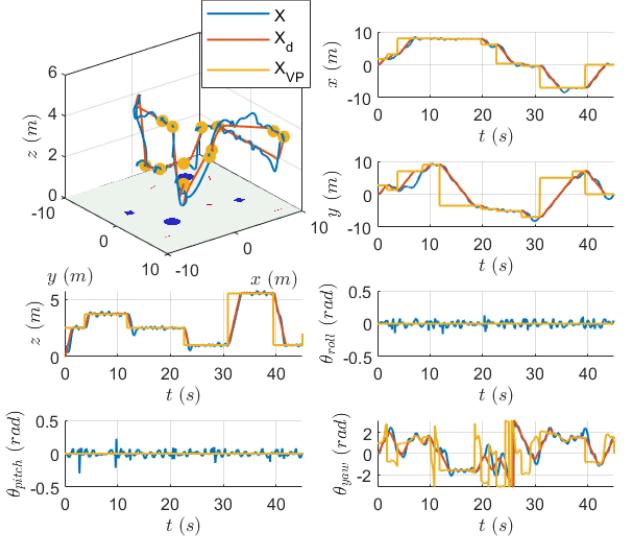}
\caption{UAV can complete a run of the circuit in \SI{45}{\second} when the feedback bandwidth is limited at \SI{10}{\hertz}, and the tracking accuracy is reduced.}
\label{fig:UAVResultsZOH}
\end{figure}

The UAV maximum linear accelerations are $20~\si{\meter~\second^{-2}}$ along the $z$-axis and $7~\si{\meter~\second^{-2}}$ along the $x$-axis and $y$-axis. The maximum angular accelerations are $1~\si{\radian~\second^{-2}}$ in all the directions. The lap time is about \SI{40}{\second} with and without the obstacles. The trajectories are smooth in both trials and respect the physical limitation of the UAV. All the simulations show how the UAV uses the interaction between the bubble and the gates to adjust the planning and succeed in the task, and the controller has a negligible error in the final state. \autoref{fig:UAVResultsZOH} shows that reducing the feedback bandwidth worsen the tracking accuracy, which is also observed in the AMR simulation. 

\subsubsection{Scalability Test}
The swarm experiment has been designed to test the scalability of the proposed method to coordinate movements of multiple independent systems. The scenario used for this is a prey-predator scenario and it has 66 degrees of freedom. This simulation is designed to test that the computational cost follows a linear scaling rather than the exponential, which is typical of optimisation algorithms due to the curse of dimensionality \cite{betts2010practical}.

The preys have the same characteristics of the UAV used for the other simulation. The predator has inertia, and maximum torques and forces are scaled by a factor of 10. The spherical bubble is increased to \SI{1.5}{\meter}. The predator and prey are randomly assigned two random independent sets of via-points in the flying space, and their desired yaw is aligned with the vector connecting their position to the assigned via-point. The simulation also tests a change of formation in the preys that start in two separate swarms and reuniting after \SI{30}{\second}. We also performed a quantitative analysis of the problem scalability by comparing the execution times required to simulate 1 and 11 UAV for the \SI{100}{\second}. Each scenario is run for 20 times and the time recorded using the Matlab Profiler. Thus, the data describe the time required for computing both the proposed architecture and the forward dynamics of the entire simulation on a single tread. It shall be noted that the physics' simulation accounts for most of the computational cost, and it is not needed on a robot. 

\begin{figure}[!tb]
\centering
\includegraphics[width=\columnwidth,trim=0cm 0.5cm 0cm -0.2cm,]{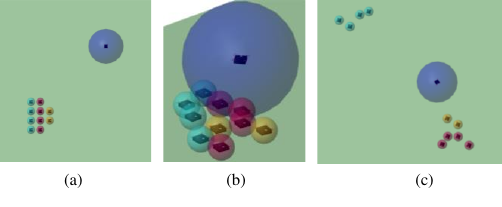}
\caption{(a) Starting condition (b) Example of real-time interactions between multiple dynamic agents (c) The formation can be managed assigning different via-points. }
\label{fig:UAVPredator}
\end{figure}

The results show the proposed method can synchronise multiple agents just acting on the assigned via-points. Furthermore, the simulation highlights how the systems can use the bubble to coordinate their manoeuvring and generate emergent synchronised behaviours (\autoref{fig:UAVPredator}). The simulation time comparison revealed a scaling factor of 9.7 between the two analysed cases. The total computational time for running the 20 simulations are \SI{2141.280}{\second} for the 11 UAV, and \SI{221.982}{\second} for the single drone, which implies that the average simulation times for a \SI{1}{\milli\second} step are $1.1\pm0.1$ \si{\milli\second} and $0.11 \pm 0.01$ \si{\milli\second}, respectively. These results also open the possibility of implementing this architecture into recursive optimisation algorithms, but it will require the implementation into a faster simulator and/or leveraging multi-threading to be computed online.

The architecture scales very well with the problem dimensionality thanks to the FIC's ability to connect multiple independent controllers in parallel which can synchronise their behaviour using the "haptic interactions"~\cite{tiseo2020}. To reproduce a comparable result with an optimisation algorithm will require an accurate dynamic model of all the world dynamics and access to its full state vector (i.e., position, velocity, acceleration, and force/torques) to ensure the numerical stability of the optimisation problem.

\subsection{Hardware experiments using Ada500}
The Ada500 is a velocity controlled robot, thus we needed to modified the planning architecture accordingly to implement an MPC to compute the next desired state for the region of attraction. The velocity control signal for the robot is obtained integrating the forward dynamics equation:
\begin{equation}
    \label{AdaDirectDynamics}
    \ddot{X}=TM^{-1}(W-W_\text{ext})
\end{equation}
where $T$ is the transformation matrix between world and robot frame, $W$ is the output wrench of the FIC$_\text{F}$ and $W_\text{ext}$ is the wrench generated by the repulsive potential field that describes the desire to avoid the obstacle. $W_\text{ext}$ is derived from a circular and rectangular potential fields surrounding the robot. The controller on the robot is running at \SI{500}{\hertz}. 

We validate experimentally that the architecture can be deployed in velocity controlled system, generating smooth adaptive trajectories based on the feedback from the motion capture without relying on numerical optimisation. During the experiment, the robot and objects movements are tracked using a 16-camera Vicon Motion Capture System at \SI{100}{\hertz}, up-sampled to \SI{500}{\hertz} to match the controller frequency. The objects are both being moved by a researcher. We remote control a mobile robot with a joystick to simulate an interaction with another robot that our system can only detect using local sensors. We use a circular field with a radius of $d_0=\SI{1.5}{\meter}$ while interacting with the researcher, and a square field when interacting with the robot.

We evaluated four tasks (see \autoref{fig:ExpSetups}) in our experiment: \textit{Free}: the robot moves between two via-points without encountering any obstacle. \textit{Push}: the robot is left static and the obstacle is used to push it away from its desired position and then removed to allow the robot to return to its desired state. \textit{Interpose}: the obstacle is encountered during the motion altering the robot longitudinal movement. The obstacle is subsequently removed to allow the robot to complete the motion. \textit{Nudge}: the obstacle is introduced laterally to the robot trajectory deviating its path from the desired trajectory.

\begin{figure}[t]
\centering
\includegraphics[width=\columnwidth, trim=5.4cm 9.35cm 4.8cm 9.20cm, clip]{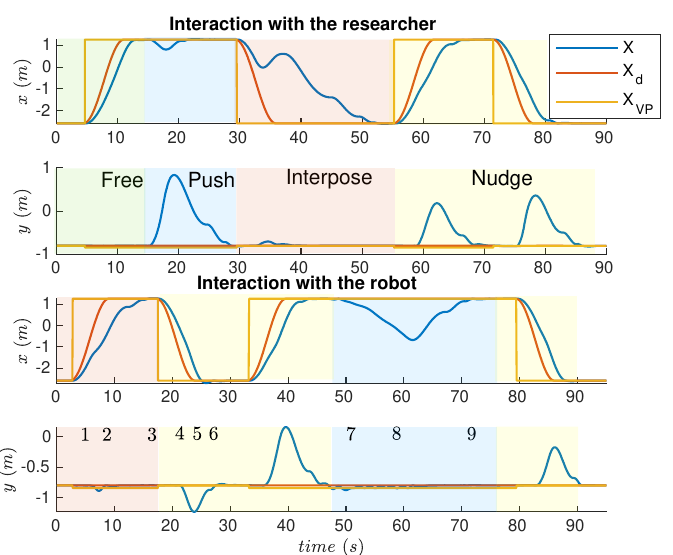}
\caption{The delay of the robot trajectory (X) with respect to the output of the elastic band (X$_\text{d}$) shows that, even if the trajectory planned by the elastic band exceeds the dynamics performance of the robot, the region of attraction properties guarantee that it is smoothly followed at the best of the hardware capabilities, as encoded in the FIC$_\text{F}$ parameters during the tuning process. The numbers reported on the last plot refer to the sequences in \autoref{fig:experimentsP}}
\label{fig:experiments}
\end{figure}

\begin{figure}[ht]
\centering
\includegraphics[width=0.9\columnwidth,trim=0cm 0cm 0cm -0.2cm,clip]{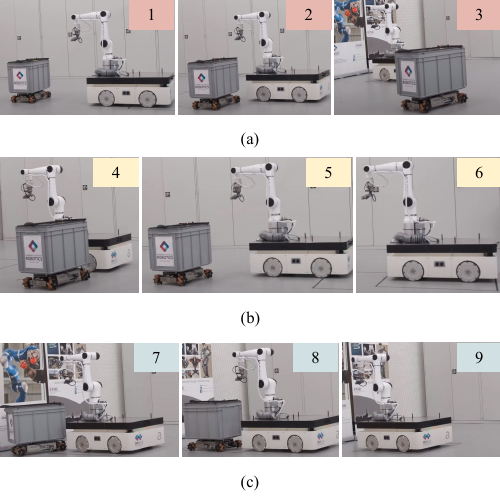}
\caption{(a) Sequences of images from the interpose trajectory, ADA slows down to give way.  (b) Sequences from the first nudge, ADA moves laterally to avoid collision before converging to the target. (c) Sequences from the push, ADA gives away its position to the other agent, and it returns in the home position once the other robot leaves.}
\label{fig:experimentsP}
\end{figure}

The experimental data show that the controller generates smooth trajectories compatible with the robot dynamics, always reaching the desired state when it was not obstructed, as shown in \autoref{fig:experiments} and \autoref{fig:experimentsP}. They also confirm that reduced feedback bandwidth does not affect the stability of the controller, enabling to control the system without state estimation on the visual feedback. The tuning of the controller also shows that even if the elastic band does not respect the mechanical limits of the physical system, the region of attraction will modulate the behaviour following the planned trajectory to the best of the hardware capabilities. The interaction with both objects shows that the robot is capable of adapting its behaviour to variable environments, diverting from the desired trajectory and recovering with smooth trajectories once obstacles are removed. It is worth noting that to guarantee obstacle avoidance requires that the bubble is proportional to the braking distance.

Comparing the planned trajectories (i.e., $X_d$) with minimum jerk trajectories with the same spatio-temporal characteristics shows that the proposed method generates a smaller maximum momentum (i.e., $M\dot{X}_\text{d-peak}$)  and  smaller peak in the system power (i.e., $M(\dot{X}_\text{d}\ddot{X}_\text{d})_\text{peak}$), making our trajectories safer and more robust than minimum jerk trajectories. For example, the values for the trajectory planned during the first experiment are: $Q_\text{max}=\SI{126.95}{\kilo\gram\meter~\second^{-2}}$ and $P_\text{max}= \SI{30.71}{\joule\per\second}$. On the other hand, the value for minimum jerk trajectory would have been: $Q_\text{max}=\SI{132.94}{\kilo\gram\meter~\second^{-2}}$ and $P_\text{max}= \SI{31.78}{\joule\per\second}$.

\section{Discussion}
The proposed architecture has been tested with two different vehicles and in a multitude of scenarios, where it had to use soft contacts and potential fields to interact with the surroundings. The simulations and hardware experiments show that the proposed architecture can plan and perform articulated motions at a low computational cost in complex scenarios. Making it extremely energy efficient and enabling the deployment in micro-controllers.

The guaranty of stability associated with the proposed method make it possible to remove the local optimisation algorithms and to decouple the global plan from local stability. Therefore, the global planning is reduced to the geometrical segmentation of a map to obtain quasi-convex subsets. These qualities are useful when dealing with unknown obstacles, which cannot be accounted in a traditional optimisation due to the lack of \textit{a priori} knowledge, but their effect ripples beyond their interaction with the system. 

Furthermore, the decoupling of the global plan from the stability of interaction enables to coordinate multiple agents without requiring complex dynamics model of the coupled dynamics, as shown in the swarm simulation  and in the experiments. This makes this architecture suitable for shared-autonomy application where the remote operator can issue via-points to multiple agents, being ensured that they are robust to environmental interaction. 

The proposed method is also robust to drift and low-bandwidth feedback in general. This capability is particularly evident in \autoref{AdaMazeR}, where the simulation shows how both the Ada robot and the UAV can complete their maze with unknown obstacles relying on a feedback bandwidth of \SI{10}{\hertz} ( \autoref{fig:UAVResultsZOH}), which is well below the performances of consumer cameras \SI{30}{\hertz} (e.g., Intel RealSense), which can be improved using sensor fusion techniques \cite{li2020}.

In summary, both simulation and experiments' results highlight the accuracy of the proposed method in reaching the desired state. They also show that the proposed method can trade-off tracking accuracy to respect internal limits and avoid obstacles. The method is also capable of autonomously recovering from large tracking errors without jeopardising stability. Nevertheless, these capabilities are reactive strategy which should be coupled with algorithms that update $X_\text{VP}$ if the distance between $X_\text{d}$ and $X$ surpasses predefined threshold levels. 

\appendix
\label{RMSE}
The tracking RMSE are:
\begin{compactenum}
\item \textbf{AMR without obstacles} \\ \noindent RMSE$_{\text{xy}}=[0.201, 0.327]~\si{\meter}$ \textemdash~ RMSE$_{\text{Y}}=0.31~\si{\radian}$
\item \textbf{AMR with obstacle} \\ \noindent RMSE$_{\text{xy}}=[0.377, 0.662]~\si{\meter}$ \textemdash~ RMSE$_{\text{Y}}= 0.31~\si{\radian}$ 
\item
\textbf{AMR  low-bandwidth} \\ \noindent
RMSE$_{\text{xy}}=[0.412, 1.96]~\si{\meter}$ \textemdash~ RMSE$_{\text{Y}}=0.70~\si{\radian}$ 

\item \textbf{UAV without obstacles} \\ \noindent
RMSE$_{\text{xyz}}=[0.152, 0.160, 0.127]~\si{\meter}$ \\ \noindent RMSE$_{\text{RPY}}=[0.001, 0.001, 0.381]~\si{\radian}$.
\item \textbf{UAV with obstacles}: \\ \noindent RMSE$_{\text{xyz}}=[0.335, 0.878, 0.119]~\si{\meter}$ \\ \noindent RMSE$_{\text{RPY}}=[0.001, 0.001, 0.2935]~\si{\radian}$ 
\item \textbf{UAV  low-bandwidth} \\
\noindent RMSE$_{\text{xyz}}=[0.410, 0.967, 0.208]~\si{\meter}$ \\
\noindent RMSE$_{\text{RPY}}=[0.036, 0.038, 0.506]~\si{\radian}$ 
\end{compactenum}
\newpage
\balance
\bibliography{Main_ICRA}
\bibliographystyle{IEEEtran}
\end{document}